# A Method for Outlier Detection Based on Cluster Analysis and Visual Expert Criteria

Juan A. Lara[1,*], David Lizcano[1], Víctor Rampérez[2], Javier Soriano[2]

[1]Madrid Open University, UDIMA, Engineering School, Carretera A6 km 38,500 – Vía de Servicio, 15 - 28400, Collado Villalba, Madrid, Spain

[2]ETS de Ingenieros Informáticos, Universidad Politécnica de Madrid, Campus de Montegancedo, 28660, Boadilla del Monte, Madrid, Spain

**Abstract**. Outlier detection is an important problem occurring in a wide range of areas. Outliers are the outcome of fraudulent behaviour, mechanical faults, human error or simply natural deviations. Many data mining applications perform outlier detection, often as a preliminary step in order to filter out outliers and build more representative models.

In this paper, we propose an outlier detection method based on a clustering process. The aim behind the proposal outlined in this paper is to overcome the specificity of many existing outlier detection techniques that fail to take into account the inherent dispersion of domain objects. The outlier detection method is based on four criteria designed to represent how human beings (experts in each domain) visually identify outliers within a set of objects after analysing the clusters. This has an advantage over other clustering-based outlier detection techniques that are founded on a purely numerical analysis of clusters.

Our proposal has been evaluated, with satisfactory results, on data (particularly time series) from two different domains: stabilometry, a branch of medicine studying balance-related functions in human beings; and electroencephalography (EEG), a neurological exploration used to diagnose nervous system disorders.

To validate the proposed method, we studied method outlier detection and efficiency in terms of runtime. The results of regression analyses confirm that our proposal is useful for detecting outlier data in different domains, with a false positive rate of less than 2% and a reliability greater than 99%.

## 1. Introduction

The analysis of large data volumes to discover new knowledge is a major challenge in the field of computer science. The discovery of useful, implicit and previously unknown knowledge from large volumes of data is a process referred to as knowledge discovery in databases (KDD). The KDD process ranges from data comprehension and preparation to results interpretation and use. Data mining is a stage within the KDD process examining the data and applying a set of techniques and tools to discover useful hidden information (Fayyad, Piatetsky-Shapiro & Smyth, 1996). There is currently a wide variety of data mining techniques for solving different problem types like, for example, classification, clustering, regression or association rule search.

Before applying data mining techniques, it is important to check data quality in order to assure that the resulting models are representative. At this stage, outlier location and filtering is regarded as a fundamental data processing task, as failure to remove outliers could undermine the resulting models (Tan, Steinbach & Kumar, 2006).

Any object in a set is considered to be an outlier if it can be classed as having features that are significantly different from the other objects within the set (Tan et al., 2006). The main outlier detection issue is to determine what these features are and decide when they are significantly different from the characteristics of the other objects. Another important issue is the possibility of objects that are not really outliers being labelled as such (false positives). Figure 1 illustrates two examples of outliers. The first example shows the average score achieved by a series of basketball players depending on their respective years of experience. It shows that one outlier is outside the normality ranges established for this domain. On the other hand, the second example represents temperature measurement inside an electronic system over a period of time, where, towards the end of the series, there are two outliers with a clearly higher score than the others.

---

[*] Corresponding author: juanalfonso.lara@udima.es

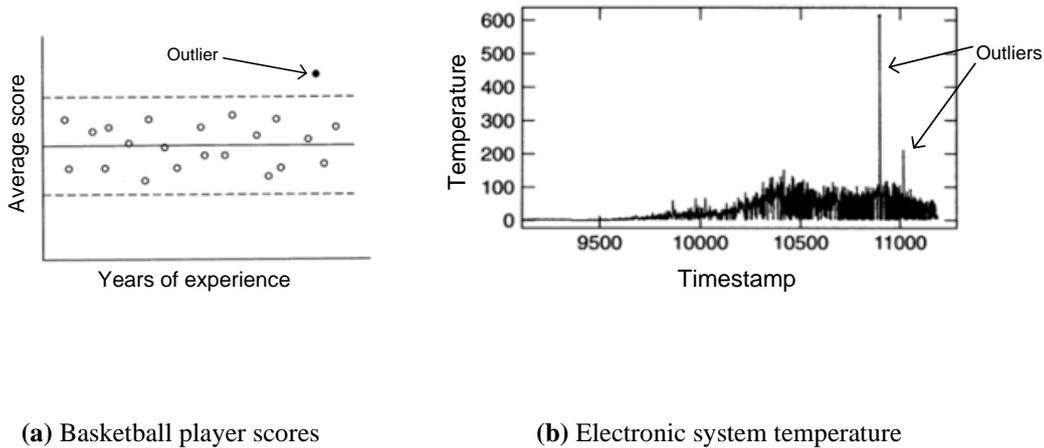

**(a)** Basketball player scores          **(b)** Electronic system temperature

**Figure 1**. Examples of outliers.

There are many domains where the focus is on outliers rather than non-outliers. Fraud detection at companies offering credit card-related services is a very clear example, where the identification of outliers is perhaps the most important data analysis issue. Outlier identification is also used in wide-ranging domains, for example, for detecting computer network attacks, outbreaks of epidemics, unidentified objects in space, tumours or abnormalities in mammograms, etc.

The literature abounds with papers on outlier detection in object populations (Chakraborty, Narayanan & Ghosh, 2019; Knorr & Ng, 1999; Ramaswamy et al., 2000; Torgo, 2007; Breunig et al., 2000). They have all undeniably led to an advance in the field and separately contribute a number of positive points. However, they also have limitations: they fail to account for there being different dataset densities, they require the specification of a value for the number of clusters on which the analysis will be based or of the number of neighbours that will be used to calculate the object outlier score.

## 1.1. Contribution

This article aims to propose a visual method based on expert criteria for detecting outliers within a dataset. The proposed method aims to serve as a data processing tool within the KDD process in order to prevent outliers from distorting the resulting models.

The aim behind the proposal outlined in this paper is to overcome the specificity of many existing outlier detection techniques that do not take into account the inherent dispersion of domain objects or that will not work properly without the definition of rather a lot of parameters. Our proposal aims to do away with this limitation and propose an automated mechanism for detecting population outliers, defined taking into account expert knowledge. To do this, first, we apply a method of agglomerative hierarchical clustering that uses the object similarity matrix to build clusters as input for the outlier identification phase. Data segmentation is a fairly common practice, put forward some time ago by Agrawal, Gehrke, Gunopulos and Raghavan (1998), for achieving a better understanding of the data and facilitating other data mining tasks. Second, we propose an outlier detection technique that aims to account for human (expert) reasoning with respect to the identification of outliers within a set. This expert-based reasoning could be summarized as follows:
  a) Objects that are more isolated, that is, have fewer neighbouring objects, are more likely to be outliers.
  b) Objects that are less like the majority of objects are more likely to be outliers.
  c) Objects that are not really outliers could be considered as such in some domains where objects are highly dispersed, making outlier detection a tricky issue.

To validate this proposal, we carried out thorough experiments with time series data from two medical domains: stabilometry and electroencephalography. In this research, the objects under analysis were represented by time series, although the proposed method is applicable to other data representations. The stabilometry dataset was composed of 6480 times series containing 4000 timestamps, whereas the electroencephalography dataset was composed of 200 time series with 1000 timestamps. We conducted studies in both domains to detect the false positive and false negative rates, and an ANCOVA to study the asymptotic behaviour of the method. We also conducted method efficiency studies analysing its computational complexity and runtimes.

## 1.2. Outline

This paper is structured as follows. Section 2 gives an overview of the major outlier detection techniques, especially techniques that include a preliminary clustering process. Section 3 describes the proposed outlier detection technique run on input from a preliminary hierarchical clustering process. Section 4 presents the results of applying the proposed method on data from the stabilometry domain, a branch of medicine studying human balance, and electroencephalography (EEG), which is a neurological exploration used to diagnose nervous system disorders. We discuss the results below in Section 5. Finally, Section 6 outlines the conclusions and future work concerning the proposal.

## 2. Related Work

A large number of outlier detection techniques have been developed over the last few decades (Agrawal & Agrawal, 2015; Hodge & Austin, 2004; Domingues, Filippone, Michiardi & Zouaoui, 2018; Ernst & Haesbroeck, 2017; Chakraborty, Narayanan & Ghosh, 2019). Many outlier detection techniques are based on the idea of finding objects that are too far removed from (or, in other words, not very like) the majority (Knorr & Ng, 1998; Knorr & Ng, 1999).

There are criticisms of this type of approaches. For example, Aggarwal and Yu (2001) suggest that they might not work properly for dispersed high dimensional data. It is true that many domains contain a lot of different clusters of objects, some of which may be unlike each other and, nonetheless, not be outliers. Consider a rugby team, for instance. A rugby team includes many, quite different clusters of players (a cluster of corpulent and ungainly players, other very athletic and nimble individuals, etc.). Each group is very different from the others, but, even so, neither should be considered as outliers. For this reason, some researchers propose outlier detection techniques based on first clustering the objects and then analysing the clusters to find outliers inside the clusters. Like our proposal, these techniques usually include some semantics based on expert knowledge, necessary, in our view, to correctly identify outliers especially in domains where data are usually disperse.

Outlier detection techniques are usually divided into four major groups as discussed below.

### 2.1. Proximity-based outlier detection

There are a great many proximity-based approaches. From the proximity viewpoint, an object is considered an outlier if it is not close to the other objects (Ren, Rahal & Perrizo, 2004; Kollios, Gunopulos, Koudas & Berchtold, 2003).

This is a relatively easy approach to apply, as it is easy to determine a proximity measure. One of the most commonly adopted measures is k-nearest neighbours, that is, the score of an object is defined as the distance of its $k$ nearest neighbours. Ramaswamy, Rastogi and Shim (2000) proposed an outlier detection method based on a neighbourhood measure, listing the objects based on this measure and identifying $t$ objects that are at the top of that list as outliers, where $t$ is a parameter entered by the user.

Techniques like these are highly dependent on the value of $k$. If a very high value of $k$ is defined, a small set of neighbouring objects could be regarded as a set of outliers if the other sets are very large and not near to the smaller set.

Although very simple to apply, the algorithms implementing this approach are computationally very costly due to the large number of proximity calculations required (Hodge & Austin, 2004). Additionally, the value of $k$ can be hard to select, especially if the data are not well-known. Neither would these techniques be suitable for wide-ranging datasets with different data densities, as there is a risk of objects being identified as outliers even if they are not.

Therefore, the main weaknesses of the above proposals are their high computational cost and the need to define the size of the neighbourhood to be taken into account in the detection process. The method that we propose aims to solve both these issues.

Another type of technique applies an equivalent approach based on the distance of each object from the others instead of the strict neighbourhood concept. One such approach was proposed by Agrawal et al. (1998), where an optimized outlier detection algorithm is used to detect outliers in disk-resident high-dimensional databases. Angiulli, Basta & Pizzuti (2006) describe another similar proposal putting forward a mechanism for predicting outliers based on the creation of a model composed of a subset of the original data known as *outlier detection solving set*. This proposal achieved good results in terms of precision and performance (runtime).

To some extent, the first steps of our proposal are similar to proximity- or distance-based techniques, as the algorithm runs on a pairwise similarity matrix between objects. We consider the use of the proximity/distance idea to be quite intuitive for detecting outliers. Therefore, we decided to include this idea in our proposal in order to improve the existing approaches and achieve an efficient method so that users do not have to define a domain-specific value for $k$.

**2.2. Clustering-based outlier detection**

Clustering-based outlier detection techniques first cluster the objects. They then analyse the clusters and, on their basis, possibly determine outliers. Data segmentation is an important source of information that can be very useful for different tasks. In particular, knowledge of the object groups is very useful for detecting outliers as they tend to be in separate groups (Jiang, Tseng, & Su, 2001).

There are many proposals that use this approach. The technique proposed by Loureiro, Torgo and Soares (Loureiro, Torgo & Soares, 2004; Torgo, 2007; Torgo, Pereira & Soares, 2009) is, like our proposal, based on the analysis of clusters output after applying a bottom-up hierarchical clustering process. In fact, this technique considers any objects that are found in clusters of a size less than $t$ to be outliers, where $t$ is a threshold value entered by the user. As far as we are concerned, this point is not enough to determine whether an object is an outlier, as, in many domains, dispersed and even isolated data are not necessarily outliers. Additionally, the end user is asked to enter the parameters, such as, for example, the number of clusters. With a view to the application of this proposal to real domains, it is inconceivable that an expert (for example, a doctor) will be able to provide these values. To circumvent this problem, our proposal reduces the number of parameters to be entered by the user for the purpose of (semi-) automation. Our proposal also considers the inherent data dispersion in each domain to prevent, wherever possible, the detection of false positives (non-outliers identified as outliers).

Wang and Chiang (2008) propose the use of SVC (support vector clustering) to detect outliers. This parameterless algorithm, which automatically determines the number of clusters, has grown in importance over recent years because it is fast and robust to noise. In this algorithm, the points are transformed to a larger-sized system, using polynomial, Gaussian or sigmoidal functions. The idea is to search for the smallest sphere containing the images of the original points. The contours of these spheres are interpreted as the cluster boundaries. The boundaries of the delimited clusters are then analysed to detect outliers, as they, if any, are usually located on the edges of the clusters.

Yang and Huang (2008) propose a spectral clustering method for outlier detection. Based on the similarity matrix between each pair of objects, this proposal builds a new matrix using the number of near neighbours of each object. They propose an algorithm that applies the k-means algorithm to the above matrix to cluster a number of clusters $k$. The resulting clusters are then divided into small and large clusters according to specified thresholds. An outlier coefficient is then calculated for each object depending on whether it belongs to a small or large cluster. Finally, the objects whose outlier coefficient is greater than an established threshold are selected as outliers.

There are some other proposals, such as the method proposed by Yoon, Kwon and Bae (2007), setting out supervised outlier detection methods based on the k-means clustering algorithm. They divide the outliers into two groups: external outliers and internal outliers. To detect outliers, they propose the use of the cubic clustering criterion (CCC) (Sarle, 1987) to automatically select $k$ and run the k-means algorithm. The domain expert uses the output clusters to determine which objects are outliers. The outliers are then removed from the set of objects. This process is repeated as many times as it takes to remove all outliers.

Finally, Stefatos and Hamza (2007) propose an outlier detection method within complex data sets, using the concepts of hierarchical clustering and principal component analysis (PCA) (Jolliffe, 1986). In this case, the PCA technique is used to transform the initial observations (many) to a (much smaller) set of new uncorrelated observations. To do this, an eigenvector matrix is built by selecting the first principal components. Finally, the outliers are detected based on a clustering process, which, after dimensionality reduction, is much more straightforward.

Like the described methods, our proposal is based on a clustering process as this is useful for establishing groups of objects based on which it is possible to run a detailed analysis in search of outliers. However, clustering-based techniques require the establishment of a parameter k (number of clusters). Our proposal can determine the above value of k without user intervention, which is one of the key advantages of our method with respect to the clustering proposals in the literature. Additionally, our proposal does not need to establish the cluster size thresholds in order to label the outlier groups, as we use the concept of the normal distribution to automate this process.

## 2.3. Density-based outlier detection

Another type of outlier detection methods are based on the density concept. In this case, an object is considered as an outlier when it is found in areas where the object density is low, that is, it is isolated and there are not many surrounding objects. Objects are labelled as outliers by calculating the inverse of the surrounding density. The technique is very similar to proximity-based outlier detection. The key difference is the manner in which the density of a point is established, as density is generally calculated from the proximity of one object to the other objects. For example, the DBSCAN algorithm (Ester, Kriegel, Sander & Xu, 1996) regards the density surrounding an object as equal to the objects that are within a specified radius of the object.

The major challenge facing algorithms that use this approach is to determine the proximity radius. If a very low value is selected, then the density of many non-outlier objects (on the edges of a dense region) will be low, and they will therefore score high as outliers. On the other hand, if the radius value is very high, then many objects will not be considered as outliers as they have a very similar density to that of the non-outliers.

The main proposal within this group of techniques, known as local outlier factor (LOF), was reported by Breunig, Kriegel, Ng and Sander (2000), establishing the use of the relative density concept. The above proposal addresses the calculation of an outlier factor for each object. This factor is calculated based on the relative density of an object with respect to its neighbours, such that an object that has a substantially lower density than its neighbours will have a higher outlier factor.

Figure 2 shows three objects *A*, *B* and *C*. Even though it is in a region with low object density, object *A* should not be considered as an outlier, as the density of its neighbour is very similar. However, object *B*, which is in a higher density region, should be considered as an outlier, as the density of its neighbours is clearly higher. Finally, object *C* is identified as an outlier as it is completely isolated.

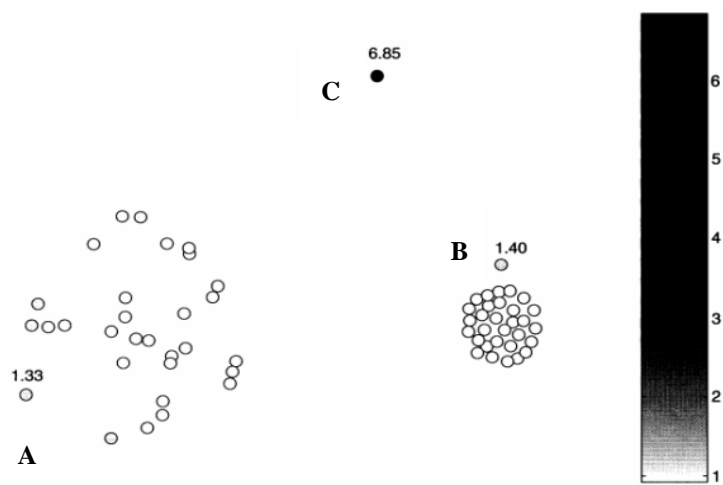

**Figure 2.** Outlier factor of three points (*A*, *B* and *C*) based on the concept of relative density.

Despite the improvements introduced by Breunig et al. (2000), we believe that, just like proximity-based techniques, density-based techniques have the drawback of not being able to correctly detect outlier objects if the data are represented by a lot of sets with different densities. This is in fact one of the major handicaps of density-based techniques, which also require users to define parameters like the radius to be taken into account to calculate the density. As mentioned above, the users of our proposal do not have to define this type of parameters, which are often outside their field of expertise.

## 2.4. Outlier detection based on statistical approaches

Outlier detection techniques based on statistical approaches are founded on the use of previously created data-based statistical models that are then used to evaluate the probability of an object fitting the respective model. If the above probability is very low, the object is considered an outlier.

The probability distribution model is built from data subject to the definition of a number of parameters. For example, if a Gaussian distribution is adopted, it will be necessary to define the mean and standard deviation in order to calculate the probability of an object being in the distribution. Other possible examples are the Poisson or the binomial data distributions. Figure 3 shows an example of a data distribution for a single attribute whose mean is 0 and standard deviation is 1. In this case, it is necessary to establish the limits (distance of the attribute to the centre of the distribution) based on which an object is considered an outlier. For example, if the minimum probability value is defined as 4.55%, then values of $x$ less than –2 and greater than 2 will be considered outliers.

For outlier detection, it is of vital importance to establish which distribution best represents the data, as if the selected model is not suitable, an object may be wrongly identified as an outlier (false positive) or not be identified as an outlier when it really is (false negative). This type of statistical approach is very effective when the data are fairly well-known. The major limitation of this type of proposals is that, although many techniques are available for evaluating a single attribute, there are not many options for assessing multiple attributes.

Despite the abovementioned difficulties, particularly the fact that statistical models (which can be highly dependent on the data domain) have to be established a priori, we believe that the use of concepts related to statistical distributions has advantages for identifying outliers. In particular, the use of the Gaussian-based normality concept is, in our view, a useful and applicable idea that has likewise been adopted in our proposal.

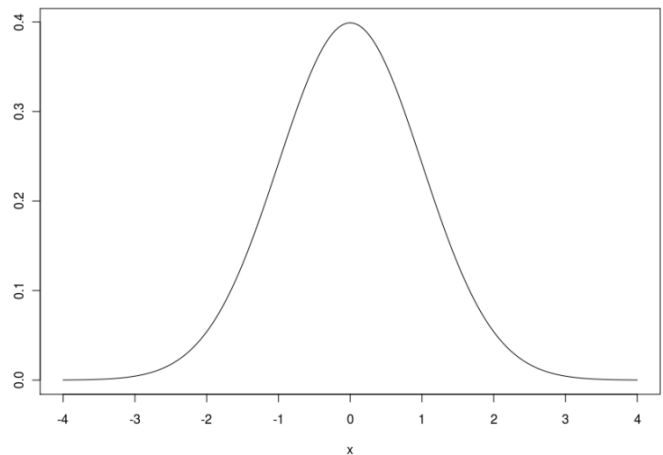

**Figure 3**. Gaussian probability distribution.

## 3. The proposed solution: a cluster analysis–based outlier detection method

Outlier detection in a dataset $D = \{O_j, 1 \leq j \leq n\}$ involves identifying any objects $O_j$ that have very different features from the other objects in the set. Section 2 reviewed the most representative state-of-the-art proposals, which contribute important and useful problem-solving ideas. However, they also have several weaknesses and shortcomings, like, for example, the need for end users to establish parameter values or the fact that they are not generalizable to domains with different characteristics (natural dispersion, dimensionality, size, etc.).

As explained in Section 2, existing outlier detection techniques are based on wide-ranging concepts. Many of these techniques require objects to be clustered beforehand. After clustering, the clusters are analysed to identify the outliers. The proposal reported here adopts this approach, which it combines with concepts borrowed from other outlier detection techniques, taking into account graphical and visual expert knowledge to define the inherent data dispersion in each domain. For the clustering phase, we used an agglomerative hierarchical clustering algorithm adapted to our needs. This clustering algorithm is based on the object similarity matrix. Similarity values are within the interval [0, 1], where 1 indicates that the two objects are identical and 0 means that they are completely different. The proposed outlier detection method designed to locate the outliers in a domain expert-like way is then applied to the output clusters.

### 3.1. Hierarchical clustering

Agglomerative (bottom-up) hierarchical clustering methods initially partition objects so that each object forms a single cluster. These clusters are then agglomerated stepwise so that the two clusters that are most alike are merged in each step. This process is repeated until there is a single cluster containing all the elements. This process is usually represented as a binary tree diagram, called dendrogram, where the leaf nodes represent the initial partition and the root node represents the final cluster containing all the objects.

After building the tree, we have to determine at which level to cut off the dendrogram in order to output the clusters (elements that are underneath each branch on the cut-off line will be in the same cluster).

The procedure of agglomerative hierarchical clustering method used is shown in Algorithm 1 (the first four steps show the general dendrogram construction process and the last two steps, a proposed division of the dendrogram into clusters).

**Algorithm 1. Hierarchical clustering**

- **Input:** set of objects D = {O$_j$, 1≤j≤n}; similarity matrix S = (s$_{ij}$) 1≤i≤n , 1≤j≤n, s$_{ij}$∈[0, 1]

- **Output:** set of clusters C

- **Steps:**

  1. Assign each object in D (O$_m$) to a cluster (C$_m$ ∈ C).
  2. Locate the most similar pair of clusters (C$_p$, C$_q$) and merge into a single cluster (C$_{pq}$).
  3. Calculate the similarity between the new cluster and the other clusters.
  4. Repeat Steps 2 and 3 until all the objects are in a single cluster of size *n*, outputting the final dendrogram *D*.
  5. Determine the threshold *T* that will be the baseline for the division of the dendrogram *D*.
  6. Fragment the dendrogram *D* to output clusters in C.

In Step 1, we assign each object to a cluster to form *n* clusters, each containing an object. The similarities between the clusters will be the same as the similarities between the objects in each cluster. In Step 2, we locate the most similar pair of clusters. They are then merged into a single cluster, outputting one less cluster. In Step 3, we calculate the similarity between the new cluster and the other clusters. The similarity between two clusters $C_i$ and $C_j$ will be the minimum similarity value between any object of $C_i$ and any object of $C_j$, as indicated in Equation (1). This value is stored in the respective dendrogram node.

$$\text{Sim}(C_i, C_j) = \min(\text{Sim}(O_k, O_l)), \forall O_k \in C_i, \forall O_l \in C_j \qquad (1)$$

We opted to use the minimum value between cluster similarity (known as single-linkage), but there are other alternatives, such as the maximum similarity value or the mean similarity value. The advantage of using the minimum similarity value is that it gives an idea of the greatest difference between the elements of a cluster.

Later, we repeat Steps 2 and 3 until all the objects are in a single cluster of size *n*. Step 4 of the algorithm outputs the dendrogram, where any dendrogram node stores the minimum similarity value between any pair of objects underneath that node. Therefore, the similarity stored in the top nodes will always be less than the similarity in the bottom nodes. In Step 5, we determine the threshold *T* (cut-off level) that will be the baseline for outputting the clusters from the dendrogram. This threshold represents the minimum similarity value that there should be between two objects if they are to be considered members of the same cluster. In fact, the *T* value represents a horizontal line. This line divides the dendrogram such that all the nodes that have a greater similarity value than *T* are below that line, and all the objects that hang from each branch on the cut-off line will be grouped in the same cluster. According to Yang and Huang (2008), the threshold should be located within the interval [μ - σ, μ), where *μ* and *σ* are the mean and standard deviation of the similarities in the matrix *S*. If *T* = *μ* - *σ*, then very few clusters are likely to be output. If *T* = *μ*, the number of clusters will be greater. After running a number of tests, we decided to make *T* equal to the interval midpoint, as shown in Equation (2).

$$T = \frac{2\mu - \sigma}{2} \qquad (2)$$

Finally, in Step 6, we fragment the dendrogram using the similarity threshold *T* determined in Step 5 and output the clusters. This guarantees that the similarity between any pair of objects belonging to the same cluster is equal to or greater than *T*.

Figure 4 shows an example of a dendrogram built from 30 objects whose root stores the value 0.23. This means that the minimum similarity value between any pair of objects is 0.23. The value of 0.35 in the left-hand subtree means that the minimum similarity value between any couple of objects underneath this value (specifically objects 18 to 30) is 0.35. The value of 0.28 in the right-hand subtree means that the similarity value between any pair of objects underneath this value (specifically objects 1 to 17) is 0.28. For the dendrogram in this example, the cut-off level was set at $T = 0.45$, outputting four clusters, one for each of the branches cut by the line specified by threshold $T$.

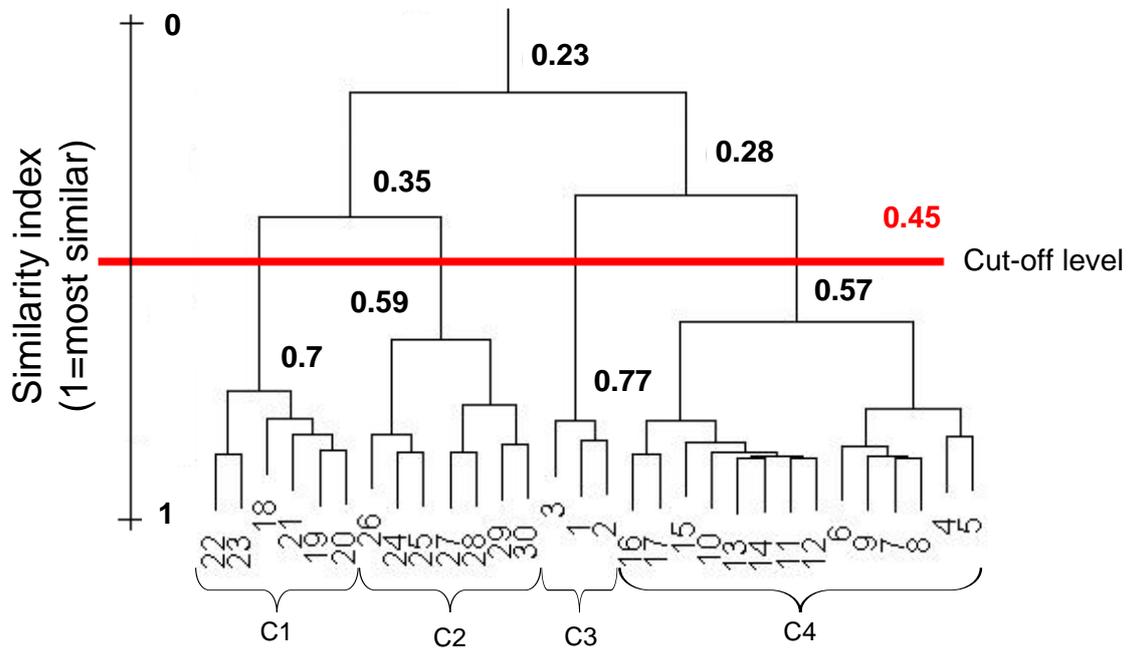

**Figure 4**. Specimen dendrogram.

The advantage of this clustering algorithm is that it is semi-automatic, that is, no user intervention is required to determine the number of clusters, which is done automatically.

### 3.2. Outlier detection

Many outlier detection techniques are based on a previous clustering process. After clustering, the clusters are analysed to determine which cluster elements can be considered outliers. Most of these techniques require user intervention to assign a value to the parameters used in the outlier detection process. Often, outlier detection is a small part of a much broader process, and the user may not be qualified to assign the correct values to these parameters.

The second problem with existing techniques is that they are usually based on purely numerical cluster analysis and do not account for how a human domain expert would detect outliers in the above clusters.

With the technique proposed in this paper, we aim to overcome these two weaknesses. On the one hand, the technique should assure that the user does not have to oversee or supervise the clustering process, and, on the other, it should reflect how an expert would analyse the clusters to identify outliers. This analysis is based on the following criteria, designed to account for how, by visually observing the clusters, experts would determine outliers:

a) There could be a cluster representing most (over 50%) of the objects. In this case, the objects in this cluster are not outliers, whereas the objects in the other clusters could be outliers.

b) The objects in the more isolated clusters (without neighbouring clusters) could be outliers. Cluster isolation can be measured as the average distance to the other clusters.

c) The smaller the percentage of objects in a cluster (with respect to all the elements), the greater the likelihood of the objects in this cluster being outliers.

d) Despite this, there may be domains where the objects tend to be very dispersed without necessarily being outliers. In such domains, the outlier detection process should be very permissive, as only objects that are extremely isolated and far removed from the others will be real outliers.

The above, originally visual and informally presented, criteria constitute an algorithm that calculates an outlier factor for each object by translating the above criteria into specific algorithmic procedures. The outlier factor accounts for outlier detection criteria a), b) and c). Finally, all the factors are analysed and the outliers are determined. In this respect, the proposed method uses the concept of outlier factor set out by Yang and Huang (2008). Our proposal differs from Yang and Huang's as to how the outlier factor is calculated based on the above criteria.

To account for criterion d), the outlier detection algorithm includes a parameter $d \in [0, 1]$, called *inherent dispersion factor*, whose value is set by the domain expert. This parameter represents the intrinsic dispersion of the domain in question. If the value of this factor is 0, it means that the objects are not expected to be very widely spread, and, therefore, any elements that are located at a distance from the majority should be considered outliers. If, on the other hand, the value is closer to 1, it means that inherent domain dispersion is expected to be high, and, therefore, the objects should be treated more permissively even if they are isolated and far removed from the others.

Our method is designed for use by highly expert users in their particular application domain. According to research in our reference domains, inherent dispersion is a known concept and, therefore, easy for experts to establish. On the other hand, note that the concept of inherent data dispersion is different to the data density concept. Inherent domain dispersion represents the expected variation in the domain data. This inherent dispersion is associated with the domain in question and not with a particular data sample of the above domain. Besides, density is a concept linked to particular data instances and could represent a specific data scatter. Therefore, there would be no equivalent in our proposal to the problem of variable densities described in density-based techniques, as inherent dispersion is a generic concept that depends on the domain and not on the specific data under analysis.

Algorithm 2 specifies the outlier detection method.

**Algorithm 2. Outlier Detection**

- **Input:** *D = {O$_j$, 1≤j≤n}*, the set of *n* objects; *C = {C$_i$, 1≤i≤k}*, the set of clusters (*k* is the number of clusters containing *n* objects); *d*, the inherent dispersion factor

- **Output:** set of outliers *OUTL*

- **Steps:**
    1. Calculate the outlier factor *OF* for each object in *D* belonging to cluster C$_i$.
    2. Calculate the mean *($\mu_{OF}$)* and standard deviation *($\sigma_{OF}$)* of the *OF*s.
    3. Set the outlier threshold *OT* using value *d*.
    4. Return outliers *OUTL*.

In Step 1, we calculate the outlier factor $OF \in (0, 1)$ for each object $O_j$ belonging to cluster $C_i$, according to Equation (3).

$$\mathrm{OF}(O_j) = \frac{\mathrm{OF}_{\#NEIGB}(O_j) + \mathrm{OF}_{LOC}(O_j)}{2} \quad (3)$$

*OF(O$_j$)* is calculated as the arithmetic mean of the other two factors, $OF_{\#NEIGB}(O_j)$ and $OF_{LOC}(O_j)$, which are calculated according to Equations (4) and (5), respectively.

$$\mathrm{OF}_{\#NEIGB}(O_j) = 1 - \frac{|C_i|}{|D|} \quad (4)$$

The $\mathrm{OF}_{\#NEIGB}(O_j)$ factor will increase as the number of objects in the cluster drops. This factor accounts for outlier detection criterion c).

$$\mathrm{OF}_{\mathrm{LOC}}\left(O_{j}\right) = \begin{cases} 1 - Sim(O_j, C_r) & si\, \exists\, C_r \\ 1 - \dfrac{\sum Sim(O_j, C_i)}{k-1} & otherwise \end{cases} \qquad (5)$$

$\mathrm{OF}_{\mathrm{LOC}}\left(O_{j}\right)$ accounts for outlier detection criteria a) and b). If there is a representative cluster of a huge majority of objects, $\mathrm{OF}_{\mathrm{LOC}}\left(O_{j}\right)$ is calculated as the similarity of object $O_j$ to the representative cluster $C_r$. If the object is a member of the representative cluster, the factor value is 0. If, on the other hand, there is no cluster representing the majority, the factor is calculated as the mean of the similarities of the object to all the clusters, except the cluster of which it is a member. In this case, the similarity of an object $O_j$ to a cluster $C_i$ ($O_j \notin C_i$) can be considered to be the maximum similarity value of object $O_j$ to any object in $C_i$. Note that the greater the value of these factors, the more likely the object is to be an outlier.

In Step 2, we calculate the mean ($\mu_{OF}$) and standard deviation ($\sigma_{OF}$) of the *OF*s. Then, in Step 3, we set the outlier threshold *OT* according to Equation (6), which will be discussed later.

$$OT = \mu_{OF} + (1 + 2d^2)\, \sigma_{OF} \qquad (6)$$

Finally, Step 4 returns as outliers any objects whose *OF* is greater than *OT* as shown in Equation (7).

$$OUTL = \{O_j \in D \mid OF(O_j) > OT,\ 1 \leq j \leq n\} \qquad (7)$$

Note that Steps 2, 3 and 4 should account for outlier detection criterion d). If the value of *d* is 0, then $OT = \mu_{OF} + \sigma_{OF}$, and, according to the Gaussian probability distribution tables, there will be a relatively high likelihood of finding an object with an *OF* value greater than *OT*. As the value of *d* increases, the likelihood of finding outliers decreases. The extreme case is when the value of *d* is 1. In this case, $OT = \mu_{OF} + 3\sigma_{OF}$, and, consequently, the likelihood of finding outliers is almost 0. Figure 5, which shows the normal or Gaussian distribution curve, illustrates this point.

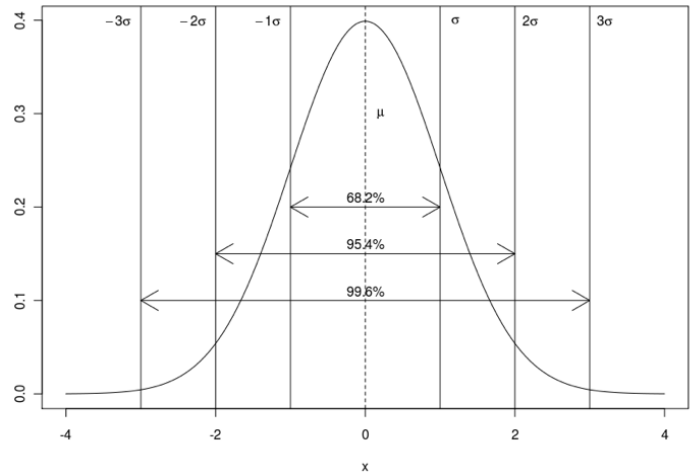

**Figure 5**. Normal distribution plot.

## 4. Validation of the proposed method

To validate the proposed method, we ran a number of tests using data from two different domains: stabilometry, a branch of medicine studying human balance, and electroencephalography, a neurological exploration used to study the electric activity generated by the brain and the nervous system.

Note that, apart from the above domains, our method is applicable to any domain where it is possible to establish a similarity measure (or distance) between each pair of objects. In the particular case of the stabilometry and electroencephalography domains, each object or individual is represented by a time series that records the balance and neurological activity, respectively, of the individual in question.

Our proposal is evaluated according to a similarity matrix between each pair of time series of each individual. To build the above matrix, we used a time series comparison method reported elsewhere (Lara, 2011; Lara, Lizcano, Pérez & Valente, 2014) and explained in Section 4.1. The data used to validate our proposal in the stabilometry and electroencephalography domains are described in Sections 4.2 and 4.3, respectively. The results are reported in Section 4.4.

### 4.1. Method used to build similarity matrices

The comparison of two time series is one of the major time series analysis problems. If a similarity measure between time series is available, it is possible to output different patterns in a set of time series, search a time series for a specified pattern, simplify a set of time series by finding similar time series, detect outlier time series, etc. In the field of economics, for example, similarity measures are applied to compare interest or inflation rates in different geographical regions or to compare time series of stock exchange values with known patterns for categorization. Most existing techniques compare one whole series with another whole series (Agrawal, Faloutsos & Swami, 1993; Chan & Fu, 1999). However, there are many problems focusing on certain regions of interest, known as events, rather than analysing the whole time series (Povinelli, 1999). This applies to areas where it is interesting to analyse short-lived events. One example is seismography, where the points of interest occur when the time series shows an earthquake, volcanic activity leading up to the earthquake or replicas. Other examples of this type of domains with time series that contain events are stabilometry and electroencephalography, our reference domains.

To determine the similarity between each pair of time series, the comparison method used identifies the events that both series have in common. The greater the number of events that two series have in common, the closer similarity will be to 1. If the series do not have any event in common, similarity will be equal to 0. The method used for this purpose was originally proposed elsewhere (Lara et al., 2014). It is outlined here less formally, using the original notation.

To determine whether an event in one time series appears in another, the event has to be characterized by means of a set of attributes and compared with the other events of the other series. To speed up this process, all the events present in the two time series are clustered. Therefore, if two events belong to the same cluster, they are similar. The goal is to find events that are members of the same cluster and belong to different time series.

Therefore, the algorithm used for extracting events common to two time series $S_A$ and $S_B$ is:

1:     **function** Similarity($S_A$,$S_B$)

2:       cluster all events $E_m$ of both time series

3:       *{events that appear in $S_A$ or in $S_B$}*

4:       **for every** cluster extracted from Step 2 **do**

5:         **while** there are events of $S_A$ and $S_B$ in the cluster **do**

6:           create all the possible event pairs ($E_{Ai}$,$E_{Bj}$)

              where $E_{Ai} \in S_A$ and $E_{Bj} \in S_B$

7:           select the event pair that

              minimizes distance($E_{Ai}$,$E_{Bj}$)

8:           *{Equation (12) describes the distance to be used.*

              *This extracts the two most alike events that are*

              *common to both series because they are in the*

              *same cluster, belong to different time series and*

              *minimize distance to each other}*

9:           delete events $E_{Ai}$ and $E_{Bj}$ from the cluster.

10:          return the pair ($E_{Ai}$,$E_{Bj}$)

11:          *{ ($E_{Ai}$,$E_{Bj}$) is an event common to both series}*

12:        **end while**

13:     **end for**

14:     return the similarity between S_A and S_B.

15:     *{similarity is calculated according to Equation(2)}*

16:     **end function**

Algorithm Step 7 extracts the events that are common to the two series, outputting a set of event pairs such that ($E_{Ai}$,$E_{Bj}$) indicates that event *i* of series $S_A$ is equal to (or similar enough to be considered the same as) event *j* of series $S_B$. Based on the above common event pairs, the similarity value between the time series is determined using Equation (8), where $E_m$ denotes each of the identified events in both series.

$$SIM(S_A, S_B) = \frac{\sum_{i,j} length\_pair((E_{Ai}, E_{Bj}))}{\sum_m length(E_m)} \quad (8)$$

In Equation (2), *length_pair* is the duration of pair ($E_{Ai}$,$E_{Bj}$), which is determined by Equation (9).

$$length\_pair((E_{Ai}, E_{Bj})) = length(E_{Ai}) + length(E_{Bj}) \quad (9)$$

Besides, *length* is the duration of a particular event $E_m$, which, as specified in Equation (10), is the absolute difference between the time at which the event ends and the time at which it starts.

$$length(E_m) = |final\_timestamp(E_m) - initial\_timestamp(E_m)| \quad (10)$$

Equation (8) should reflect the following idea: the aim is to compare the number of time series that are common to the two times series (numerator) with respect to the total number of useful time series, that is, with respect to the total length of the events (denominator). The more events there are in common, the greater the numerator will be, and the similarity value will, therefore, be higher. If there is no event in common, both the numerator and the similarity value will be 0. If all the extracted events are common, the similarity value will be 1. The denominator of this equation could exceptionally be 0. This would occur when there are no events for analysis. In this case, the value 1 will be assigned to similarity.

Equation (8), which is used to calculate the similarity between time series, does, in fact, have the same underlying idea as the Jaccard similarity coefficient. This coefficient, formulated as shown in Equation (11), measures the similarity between two sets of elements *A* and *B*, defined as the intersection between two sets (common elements, $A \cap B$) divided by the union of the same two sets (all the elements of the two sets, $A \cup B$) (Jaccard, 1908).

$$SIM\_Jaccard(A, B) = \frac{|A \cap B|}{|A \cup B|} \quad (11)$$

We looked at other measures of similarity when we proposed this method, including Sorensen's similarity index (Sorensen, 1957). However, we opted for a similarity formula based on the Jaccard index, as it is especially suited for quantitative data, is a normalized similarity measure and adopts the same idea as set out by our comparison method for identifying common elements.

We should also explain some aspects of the event clustering process (Step 2). To cluster events, it is necessary to calculate the distance between each pair of events explained under algorithm Step 2. The distance between events is also used in Step 7. In both cases, we opted for the city-block distance (Black, 2006). This distance calculates the sum of the absolute differences between each of the coordinates of two vectors.

$$d_{ij} = \sum_{k=1}^{p} |x_{ik} - x_{jk}| \quad (12)$$

In Equation (12), *i* and *j* are the vectors to be compared, and *p* is the number of coordinates (dimension). Again, we considered other distance measures. However, we finally chose the city-block distance, because it uses the mean distance per attribute during the clustering process to determine whether two elements are members of the same cluster. This mean distance per attribute is calculated straightforwardly by dividing the total distance $d_{ij}$ by the number of attributes *p*. Thus, the city-block distance saves time, since it obviates additional transformations that would make the clustering process more complex to develop and more computationally intensive.

Bottom-up hierarchical clustering techniques were used.

### 4.2. Electroencephalography

An electroencephalogram machine is a device used to create a picture of the electrical activity of the brain. It has been used for both medical diagnosis and neurobiological research, especially to study the neuronal correlates of mental activity. The essential components of an EEG machine include electrodes, amplifiers, a computer control module, and a display device. EEG machines are used for a variety of purposes. In medicine, they are used to diagnose such things as seizure disorders, head injuries, and brain tumours.

Decision making based on EEGs, which are time series of signals, often requires information regarding the signal characteristics (Wolpaw et al., 2000; Tzallas, Tsipouras & Fotiadis, 2007; Barry, Clarke & Johnstone, 2003; Kovalerchuk, Vityaev & Ruiz, 2000; Hassan & Subasi, 2017; Salma et al., 2018). Electroencephalogram analysis is a very useful technique for investigating the activity of the central nervous system. It provides information related to brain activity based on measurements of electrical recordings taken on a subject's scalp. The information output by EEGs can be analysed to make inferences and conduct studies with respect to patient health and the effective treatment of many diseases. EEG analysis has often been used to help medical doctors with their diagnostic procedures using information technology, especially whenever there are problems of differential disease diagnosis. Intelligent system methods provide the opportunity to formalize medical knowledge and standardize diagnostic procedures in specific domains of medicine for storage in computer systems.

All EEG signals considered in this research were recorded with the same 128-channel amplifier system, using an average common reference. The data were digitized at 173.61 samples per second using 12-bit resolution. Band-pass filter settings were 0.53–40 Hz (12dB/oct). Typical EEGs are illustrated in Figure 6.

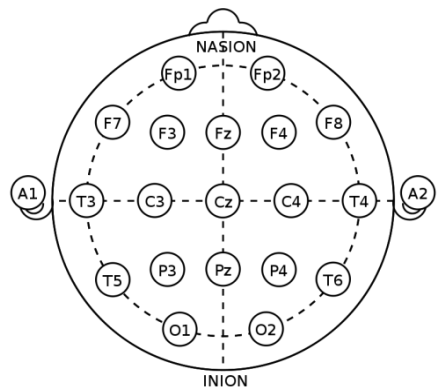

**Figure 6**. 10-20 international system of electrode placement (taken from https://en.wikipedia.org/wiki/10%E2%80%9320_system_(EEG)).

In our research, we have used the publicly available data described by Andrzejak et al. (2001), comparing the dynamic properties of brain electrical activity from different recording regions and from different physiological and pathological brain states. Using the nonlinear prediction error and an estimate of an effective correlation dimension in combination with the method of iterative amplitude adjusted surrogate data, they analyse sets of electroencephalographic EEG time series.

The complete data set consists of five sets (denoted A–E), each containing 100 single-channel EEG segments. These segments were selected and cut out from continuous multi-channel EEG recordings after visual inspection for artefacts, e.g., due to muscle activity or eye movements. Sets A and B consisted of segments taken from surface EEG recordings that were carried out on five healthy volunteers. Volunteers were relaxed in an awake-state with eyes open (A) and eyes closed (B), respectively. Sets C, D, and E were taken from the pre-surgical diagnosis EEG archive. Figure 7 shows examples of five different sets of EEG signals for different subjects.

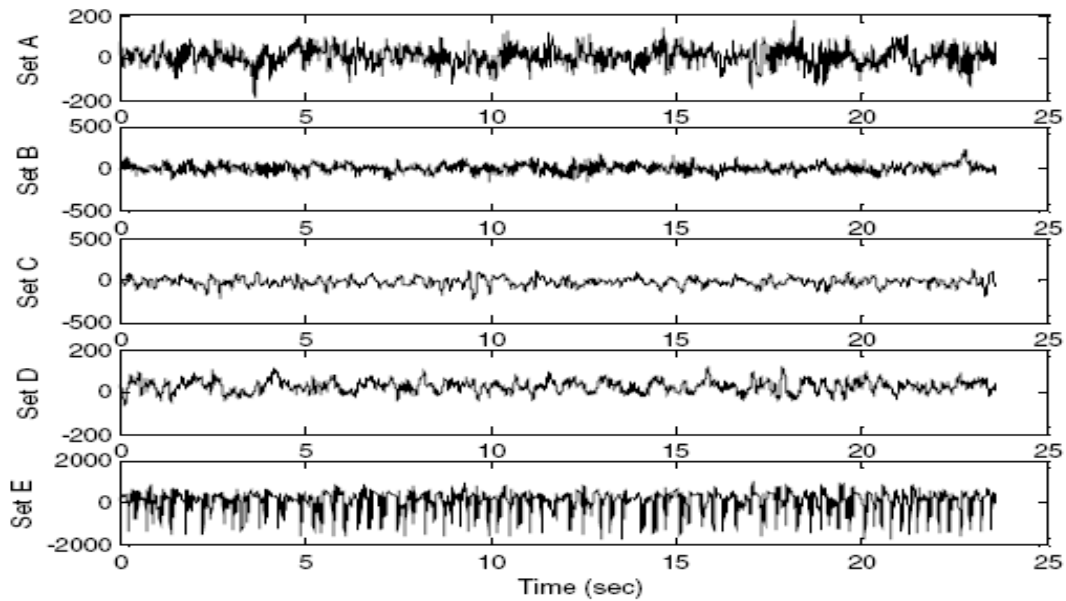

**Figure 7**. Example of five different sets of EEG signals taken from different subjects (Andrzejak et al. (2001)).

**4.3. Stabilometry**

Stabilometry is a set of techniques for analysing human postural control (Barigant, Merlet, Orfait & Tetar, 1972; Boniver, 1994; Yamamoto et al., 2018; Takada, 2019). It is also referred to as posturography, statokinesimetry and posturometry. To gather information on postural control, stabilometry uses dynamometric platforms, which are sensitive to the horizontal and vertical forces. These platforms are connected to computer systems that are capable of displaying the centre of gravity. Figure 8 shows a patient performing a test on a stabilometry platform.

Stabilometry dates back to around 1850, when Romberg (1853) conducted a number of studies to check the postural sway of individuals with their eyes open and closed, and Barany (awarded the Nobel Prize in Physiology and Medicine) described postural instability and explored the vestibulospinal function in patients with vestibular injuries (Stockwell, 1981). A series of postural control techniques were developed from the inception of stabilometry to the end of the 19th century. However, they were very painful and invasive for the patient. A second research trend then emerged that aimed to record sway by analysing the pressure exercised by the subject on a platform. This research trend led to Baron's statokinesiometer (1964), a posturographical system that was composed of four electromagnetic pressure sensors and was used widely. Baron's statokinesiometer can be considered the forerunner of current stabilometric systems.

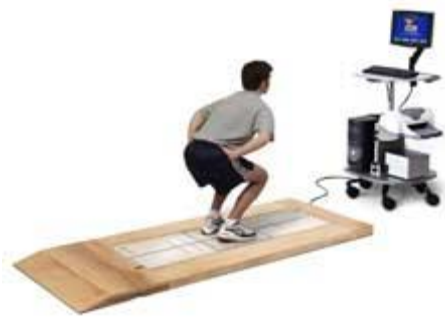

**Figure 8**. Patient performing a test on a stabilometric platform.

Postural control is a key element for understanding how able a person is to perform everyday activities. The aim of postural control is to maintain the balance and equilibrium of the body either at rest (static balance) or in motion or subject to different stimuli (dynamic balance). It has two major objectives:

- Stability, keeping the projected centre of mass within the supporting base.
- Orientation, that is, the ability to maintain an adequate relationship between the different parts of the body and between these and the surrounding environment.

To measure postural control, the patient performs a series of tests. The different tests are designed to isolate the principal sensory, motor and biomechanical components that contribute to balance with the aim of being able to evaluate the individual's ability to use these components separately or together (Sanz, 2000).

Although, in principle, stabilometry was designed merely as a postural control and balance assessment technique, it is currently regarded as a useful tool for patient diagnosis (Ronda, Galvañ, Monerris & Ballester 2002; Rama & Pérez, 2003) and for the treatment of balance-related disorders (Barona, 2003). Some examples of its use are:

- Analysis of the influence of age and sex on postural control, especially the study of imbalance in the elderly and people with motor disorders (Raiva, Wannasetta & Gulsatitporn, 2005; Lázaro et al. 2005; Nguyen, Pongchaiyakul, Center, Eisman & Nguyen, 2005; Sinaki, Brey, Hughes, Larson & Kaufman, 2005).
- Analysis of stability in patients with neurological diseases (Martín & Barona, 2007; Ronda et al., 2002).
- Study of the effect of diverse drugs on stability (Song, Chung, Wong & Yogendran, 2002).

There are two fundamental divisions with modern posturography:

a) <u>Static</u>. Static posturography uses fixed platforms to measure patient sway through the pressure exerted by their feet on the platform.

b) <u>Dynamic</u>. Dynamic posturography is based on the use of a platform placed on a support capable of moving horizontally, inclining forward or backward and rotating around an axis that is collinear with the ankles. One of the main dynamic posturography systems was developed by Nashner (Bowman & Mangham, 1989) and later studied by Black (1984; 1985).

In our research, we used a modern static posturography device: *Balance Master* commercialized by NeuroCom® International (2004). This device is composed of a metal platform that is placed on the floor and divided into two lengthwise platforms connected with each other. The metal platform is surrounded by a wooden platform whose sole mission is to stop patients tripping and falling. The patient has to stand on the metal platform to perform a number of tests. UNI, LOS and RWS are the tests that provide experts with most information (Lara, 2011).

To evaluate the method proposed in this research, we used time series for a total of 120 patients, 60 of whom had Ménière's disease, a disease that affects the inner ear and is primarily characterized by vertigo (International Statistical Classification of Diseases and Related Health Problems (ICD) H81.0), and the other 60 constituting a control group of healthy patients with the same characteristics. Vertigo-related diseases are some of the disorders most commonly studied in the field of stabilometry.

All 120 patients performed the UNI, LOS and RWS stabilometric tests to measure balance. These tests were then run under different conditions and several times according to the medical protocol. The number of time series for validation was 6480, a reasonable number taking into account that it is hard to get access to this type of private information and stabilometric tests are highly complex (provide a great deal of information but are at the same time costly to perform in terms of time and resources).

**4.4. Results**

This section reports the results of running the experiments on the above data sets. Section 4.4.1 describes the results of evaluating method quality, and Section 4.4.2 compares our method with similar proposals. Finally, Section 4.4.3 studies the complexity and efficiency of our method.

*4.4.1. RQ1: Method quality*

To evaluate the quality of the outlier detection method, we used the above sets of time series for both the electroencephalography and stabilometry domains. Table 1 summarizes the above time series, including the name of the set, the number of timestamps in the series and the number of time series in each set.

**Table 1**. Data used in experiments

| Name of set | #Time series timestamps | Number of samples |
|---|---|---|
| *EEG_Epileptic* | 1000 | 100 |
| *EEG_Healthy* | 1000 | 100 |
| *STAB_*Ménière | 4000 | 3240 |
| *STAB_Healthy* | 4000 | 3240 |

As part of the evaluation, an expert in each domain was asked to visualize the time series in question and determine which should be considered outliers. Likewise, experts were asked to establish the inherent data dispersion parameter, whose value was 0.4 for electroencephalographic data and 0.2 for stabilometric data. Our system immediately identified the outliers, which were compared with the expert decisions, as summarized in Table 2. Table 2 contains the names of the datasets, the series identifiers, the number of samples in each set, the number of outliers identified by the expert, the number of outliers located by the system, the number of false positives and false negatives output by the system.

**Table 2**. Overall results of experimentation

| Set | TimeSeries_Id | Number of samples | Expert outliers | System outliers | False positives | False negatives |
|---|---|---|---|---|---|---|
| *EEG_Epileptic* | EEG_E1..EEG_E100 | 100 | 5 | 5 | 0 (0%) | 0 (0%) |
| *EEG_Healthy* | EEG_H1..EEG_H100 | 100 | 8 | 8 | 0 (0%) | 0 (0%) |
| *STAB_*Ménière | STB_M1..STB_M3240 | 3240 | 108 | 109 | 2 (1.85%) | 1 (0.93%) |
| *STAB_Healthy* | STB_H1..STB_H3240 | 3240 | 167 | 169 | 3 (1.8%) | 1 (0.6%) |

For the statistical study of the results, we decided to divide outcomes into subsets of time series to provide more points for building a more reliable regression model reflecting the evolution of the rate of false positives depending on the number of outliers detected by the system. The first two sets were not divided as they were relatively small. The stabilometric sets were divided taking into account the stabilometric test to which each series belongs, UNI, LOS or RWS. Table 3 shows the results for the separated data.

**Table 3**. Results of the experimentation divided into subsets

| Set | Number of samples | Expert outliers | System outliers | False positives | False negatives |
|---|---|---|---|---|---|
| *EEG_Epileptic* | 100 | 5 | 5 | 0 (0%) | 0 (0%) |
| *EEG_Health* | 100 | 8 | 8 | 0 (0%) | 0 (0%) |
| *STAB_Ménière_UNI* | 720 | 4 | 4 | 0 (0%) | 0 (0%) |
| *STAB_Healthy_UNI* | 720 | 6 | 6 | 0 (0%) | 0 (0%) |
| *STAB_Ménière_LOS* | 1440 | 49 | 49 | 1 (2.04%) | 1 (2.13%) |
| *STAB_Healty_LOS* | 1440 | 105 | 107 | 2 (1.9%) | 0 (0%) |
| *STAB_Ménière_RWS* | 1080 | 55 | 56 | 1 (1.82%) | 0 (0%) |
| *STAB_Healthy_RWS* | 1080 | 56 | 56 | 1 (1.79%) | 1 (1.89%) |

The data shown in Tables 2 and 3 were used to conduct a regression analysis of the false positive variable depending on the descriptive variables of each data set. We analysed covariance to, on one hand, check the degree of dependency of this result on each of the inherent dataset characteristics and, on the other hand, find a regression model that can interrelate the analysed variable with the other variables. Thus, we first validated the analysed sample for bias, and we were then able to describe how the analysed variable evolves when the other dimensions are modified (data size, number of data outliers, etc.). The results of this study are shown in Table 4.

**Table 4.** Regression study of the false positive variable

| Adjustment coefficients: | |
|---|---|
| R (correlation coefficient) | 0.987 |
| R² (coefficient of determination) | 0.999 |
| Adjusted R². (adjusted coefficient of determination) | 0.999 |
| SCR | 0.001 |

Evaluation of the value of the information from the variables (H0 = Y=Moy(Y)):

| Source | DoF | Sum of squares | Mean square | Fisher's F | Pr > F |
|---|---|---|---|---|---|
| Model | 6 | 7.163 | 1.194 | 17624.999 | 0.006 |
| Residuals | 1 | 0.000 | 0.000 | | |
| Total | 7 | 7.163 | | | |

Model parameters:

| Parameter | Value | Standard deviation | Student's t | Pr > t | Lower bound 95 % |
|---|---|---|---|---|---|
| Intersection | -0.187 | 0.009 | -20.544 | 0.031 | -0.303 |
| Number of samples | 0.003 | 0.000 | 102.813 | 0.006 | 0.002 |
| Expert outliers | -0.018 | 0.000 | -48.036 | 0.535 | -0.023 |
| Set-EEG_Epileptic | 0.000 | - | - | - | - |
| Set-EEG_Healthy | 0.055 | 0.012 | 4.636 | 0.013 | -0.096 |
| Set-STAB_Ménière_UNI | -1.747 | 0.017 | -102.243 | 0.006 | -1.964 |
| Set-STAB_Healthy_UNI | -1.710 | 0.016 | -103.953 | 0.006 | -1.920 |
| Set-STAB_Ménière_LOS | -0.888 | 0.019 | -47.549 | 0.013 | -1.126 |
| Set-STAB_Healthy_LOS | 0.000 | - | - | - | - |
| Set-STAB_Ménière_RWS | 0.000 | - | - | - | - |
| Set-STAB_Healthy_RWS | 0.000 | - | - | - | - |

As the data volumes are constant, it is not necessary to conduct Tukey HSD or Levene tests to check for biases or possible explanations of changes of sample mean or variance.

Since the factor most correlated with the percentage of false positives is the outlier volume, the next step was to build a regression model for the false positive variable with respect to this explanatory variable. To build this model, we tested three types of adjustments: a simple linear adjustment and two parametric nonlinear adjustments based on exponential and logarithmic approximations. Table 5 shows how good each adjustment is.

**Table 5.** Goodness of the different regression adjustment types

| | Linear regression | Exponential regression | Logarithmic regression |
|---|---|---|---|
| R (correlation coefficient) | 0.881 | 0.750 | 0.971 |
| R² (coefficient of determination) | 0.775 | 0.562 | 0.943 |
| Adjusted R² (adjusted coefficient of determination) | 0.738 | 3.301 | 0.407 |
| SCR | 1.609 | 0.750 | 0.971 |

In view of the adjustment data, the regression model that best describes the result for the false positive variable is the logarithmic model ($R^2$ and adjusted $R^2$ closest to 1), as shown in the regression plots for each adjustment, illustrated in Figure 9.

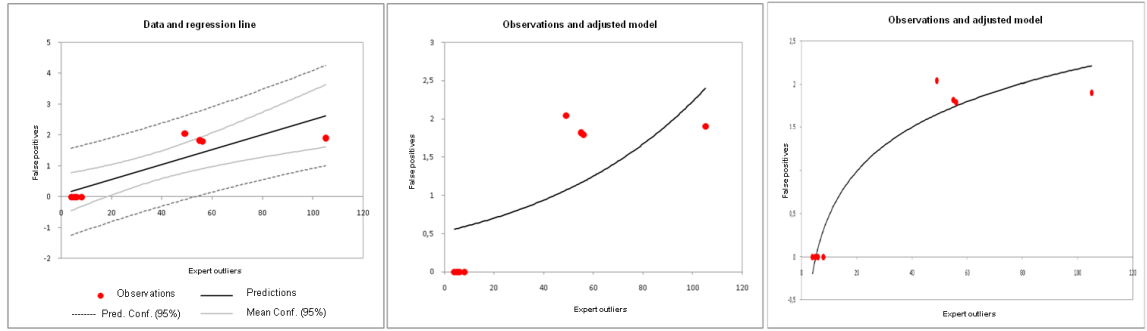

**Figure 9**. Regression plots for the analysed models.

According to the above model, the percentage of false positives can be calculated using the equation Y= 0,74*Ln(X) – 1.22.

After conducting this overall analysis of the quality of the proposed method, we ran a detailed analysis of the results for each domain. The precision of the method was analysed in depth using electroencephalographic data, as there were fewer and more manageable data in this domain, whereas effectiveness was studied in detail considering the stabilometric data, as there were a large number of samples.

With respect to the electroencephalographic data, the method determined that there were three clusters within epileptic patients, as shown in Table 6. Table 6 shows the clusters, the number of objects in each cluster, the detected outliers and the mean and standard deviation values for the outlier factor *OF*. It is clear that the value for this factor is highest for cluster $C_3$, which makes sense considering the isolation of the elements of this cluster. The value of the outlier threshold calculated using Equation 6 was 0.624, suggesting that the elements of $C_3$ are outliers according to the system, which is entirely consistent with the expert stipulations.

**Table 6**. Detailed analysis of results for the EEG_Epileptic set

| Cluster_Id | Number of objects | Detected outliers | $\mu_{OF}$ | $\sigma_{OF}$ |
|---|---|---|---|---|
| **Inherent Dispersion** = 0.4 (determined by the expert) | | | | |
| $C_1$ | 62 | 0 | 0.241 | 0.197 |
| $C_2$ | 34 | 0 | 0.573 | 0.193 |
| $C_3$ | 5 | 5 | 0.928 | 0.147 |
| *TOTAL* | **100** | **5** | **0.372** | **0.191** |
| **OF threshold** = 0.624 | | | | |

For healthy patients, the method determined that there are five clusters, as shown in Table 7, structured as above. In this case, the outlier factor was rated as 0.5777, which indicates that the elements of clusters $C_2$, $C_4$ and $C_5$ are outliers. The result for healthy patients also matched the expert verdict.

**Table 7**. Detailed analysis of results for the EEG_Healthy set

| Cluster_Id | Number of objects | Detected outliers | $\mu_{OF}$ | $\sigma_{OF}$ |
|---|---|---|---|---|
| **Inherent Dispersion** = 0.4 (determined by the expert) | | | | |
| $C_1$ | 47 | 0 | 0.326 | 0.181 |
| $C_2$ | 2 | 2 | 0.923 | 0.112 |
| $C_3$ | 45 | 0 | 0.335 | 0.178 |
| $C_4$ | 4 | 4 | 0.912 | 0.135 |
| $C_5$ | 2 | 2 | 0.931 | 0.098 |
| *TOTAL* | **100** | **8** | **0.363** | **0.162** |
| **OF threshold** = 0.577 | | | | |

With regard to the detailed analysis of the results in the stabilometry domain, note that, in this case, the discrepancies between our method and the expert were minimal, as shown in Table 1.

*4.4.2. RQ2: Comparison with other proposals*

Section 2 discussed the existing outlier detection proposals, outlining the drawbacks of each technique type, such as:

a) Specification of a number of neighbours to calculate the proximity of an object
b) Specification of a number of clusters for later analysis.
c) High computational cost.
d) Existence of different dataset densities
e) Difficulty establishing statistical models that are representative of data.

Despite the above drawbacks, the existing techniques also have major advantages and strengths. They have been adopted in our proposal with a view to improving the above techniques and providing a general-purpose and semi-automatic mechanism for identifying outliers.

Of the existing proposals, the one that is most like our proposal was reported by Torgo et al. (2009). This proposal reports a clustering mechanism that requires the user to specify the value of *k* (number of clusters). Our proposal takes the onus off the user in this respect by automatically determining the number of clusters. On the other hand, the output clusters (Torgo et al., 2009) are subject to a cluster analysis that merely takes into account their size such that outliers are identified based on the threshold parameter *t* indicating the maximum size of a cluster whose members are considered outliers. Our proposal aims to go a step further, conducting a broader analysis of the clusters considering not only their size but also their distance with respect to the other clusters and the inherent data dispersion of each domain, both of which are issues overlooked by Torgo et al. (2009). To check out the benefits of including the above two factors, we delved deeper into the analysis described in Section 4.4.1 and summarized in Tables 6 and 7, and we compared the outcomes of our method and Torgo et al.'s proposal. Focusing on the clusters of epileptic individuals (Table 6), we find that the technique described by Torgo et al. (2009) works correctly with an assignment of values greater than 5 to *t*. However, the Torgo method would not detect the five outliers for values of *t* equal to or less than 5, where there are thus five false negatives. Looking now at healthy individuals (Table 7), the method reported by Torgo et al. (2009) would work correctly when *t* is greater than 4, as it would correctly detect outliers. However, for values of less than 4, the method would not work, as it would not identify outlier individuals as outliers. There would be 4 and 5 false negatives with values of *t* less than 4. In both cases, our method is found to work at least as well as the most similar method. With the data used, our method works better as the false negative rate is much lower.

*4.4.3. RQ3: Performance*

With respect to the efficiency of the proposed method, we carried out a study to measure the execution time with respect to the number of samples. The datasets used are not large enough to be able to carry out a proper efficiency study. Therefore, we have taken the stabilometric data of healthy and epileptic patients, and we have randomly created aggregate sets of input data which were fed into the algorithm. The size of these test datasets ranged from 0 to 1000000 data, a large enough size to measure method complexity and test its scalability with large input datasets. The experiments were conducted on a machine with an Intel Core i3 2.4 GHz processor with 4GB of RAM, running Ubuntu v16.04. The evolution of runtime depending on the number of input data is shown in Figure 10.

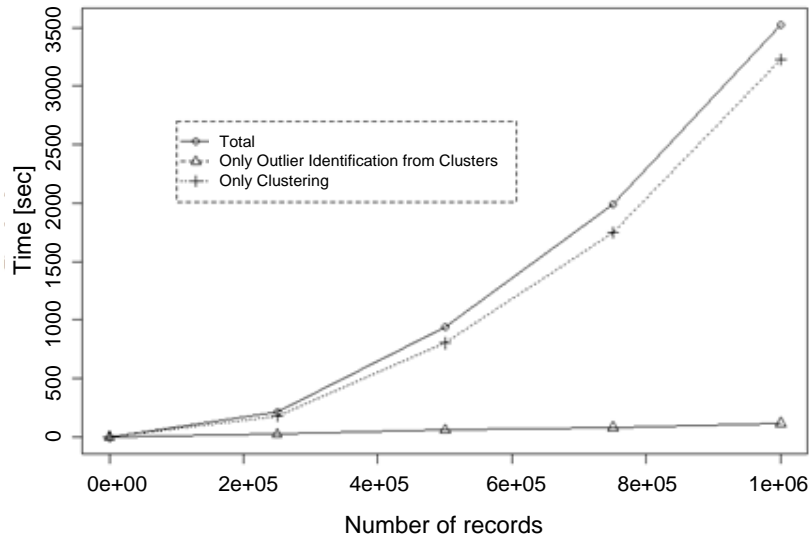

**Figure 10**. Runtime evolution depending on the number of input samples.

Figure 6 shows the clustering time and clustering-based outlier detection time separately and globally. Most of the time (with values ranging from 84% to 91%) is consumed by clustering, whereas only a small part (from 9% to 16%) is used for outlier detection. For example, the clustering time for 200000 records was 209 seconds, whereas the outlier detection time was 39 seconds. More computational details are provided below.

Looking at the different points of the graph in Figure 10, we also conclude that the complexity of our method is quadratic, as expected. On the one hand, the complexity of the bottom-up hierarchical clustering algorithm is $O(N^2)$, whereas the cluster-based outlier identification process is, at worst, $O(N)$. The worst case will occur when there are the same number of clusters and objects. The overall complexity of the method is, therefore, the most restrictive of the above two, that is, $O(N^2)$. Proposals like ours, such as the one described by Torgo et al. (2009), are similarly complex, as they are based on a bottom-up hierarchical clustering process with similar characteristics to ours in terms of efficiency.

## 5. Discussion

As shown above, thorough experiments have been conducted to test the quality of the proposed outlier detection method, using a rather large set of quite complex real data.

The system validation focused on the rate of false positives detected by our system (Table 2), as this is considered to be the key estimator of the quality of outlier identification methods. Only five false positives were detected in all of the experiments conducted. This is a false positive rate of less than 2%. As far as false negatives are concerned, the method also proved to perform excellently, as, of the four conducted experiments, only two detected a false negative, which is equivalent to an error rate of less than 1%. Rates lower than 10% are considered to be acceptable in this type of approaches.

Note that, both the false negative and the false positive rates in the set of electroencephalographic data were 0. Therefore, the method correctly identified the exact number of events detected by experts. It is true that the number of outliers was much smaller in the electroencephalographic case study. This point would appear to suggest that the method works especially well in sensitive domains where there are few outliers. Outlier detection in such domains is even more crucial, because, as there very few outliers, any error will drive up the overall percentage error enormously.

In the stabilometric domain, where there are more outliers, we conducted a detailed analysis by exercise type and found that the errors were more or less uniformly distributed by exercise type. This, again, is a positive sign, suggesting that the method performs well in different domains (Table 3).

On the other hand, the ANCOVA that we conducted (Table 4) indicates that, for a 95% confidence interval, the false positives variable has a 98.7% correlation with the set of qualitative and quantitative variables under study, namely, sample size, analysed medical domain and number of outliers identified by the expert. This means that the ANCOVA model can predict the percentage of false positives for each set of study variables with a reliability of 99.9% (value indicated by $R^2$). This suggests first and foremost that the proposed model performs predictably and deterministically. Fisher's F is also very high for the number of degrees of freedom, which means that the result of outlier detection will be predictable and depend on the analysed domain and the outliers identified by the expert. This assures that the system behaves like an expert in terms of outlier detection. The probabilities $Pr > t$ suggest that the factor that most affects the percentage of false positives is the number of outliers identified by the expert (this is the highest-scoring factor at 0.535). Accordingly, we can state that statistically the data sample is not biased, and the model is, in principle, independent of the analysed domain. This is a desirable feature for any outlier detection mechanism.

We also conducted a regression analysis (Figure 9) to check the method's asymptotic performance when certain parameters were changed. The resulting regression equation suggests that the percentage error will not be much affected, even if the number of identified outliers increases. In fact, it lies below 10% in the vast majority of cases, and this is an acceptable threshold for this type of methods. Only in samples where the number of outliers is greater than 3.8 million ($e^{15.1621}$) is the proposed method likely to fail, producing more than 10% false positives.

Finally, we carried out an introspective study of the analysed objects (Tables 6 and 7). We analysed the discrepancies between our method and reality and reached the conclusion that they were due, in all cases, to the assigned outlier factor value being very close to the threshold. This suggests that the wrongly identified individuals were misclassified by a very close margin. The studies carried out suggest that the weight assigned to the $OF_{LOC}$ measure in the equation for calculating the outlier factor could be too large, as the results improved if the weight of this factor was reduced. Despite the few false positives, we are considering conducting a study on how to optimize the weight adjustment process for each criterion used to identify outliers for future research.

To compare our method with other techniques, we also conducted a more qualitative study, described in Section 4.4.2. This small-scale study aims to illustrate the drawback of having to establish the correct parameter values in order to identify outliers, especially when they are outside the expertise and beyond the control of the domain experts. A correct initialization of the above parameters can lead, as we have seen, to poor quality outlier detection. Therefore, our method was designed with the aim of minimizing the use of the above parameters. In fact, only one parameter —inherent data dispersion—, has to be entered. This is a parameter with which domain experts are acquainted and find easy to establish.

On the other hand, based on the data from the experiments measuring method efficiency (Figure 10), we find that the clustering process will determine method scalability, although it is true, as mentioned later, that this is a process that is used routinely by businesses and institutions worldwide despite its cost. Additionally, the method performed acceptably even a number (one million) of objects that is much larger than usually occurs in typical domains (that is, a rather theoretical value) within an acceptable time period of less than one hour (57 minutes). Perhaps other more efficient proposals would be recommendable in critical real-time domains of this or similar size.

One last issue worth considering within this section is the representativeness of the studies reported here and the applicability of the method in other academic, industrial and commercial spheres. In this respect, note that we believe that the two analysed domains are sufficient to show that the proposed method is useful and efficient. Although we examine two branches of medicine, both the characteristics of the data (size, dimensionality, natural dispersion) and the experts involved in the experiments are very different. They are, in actual fact, completely unconnected branches of medicine (one related to balance and the other to neurology).

Obviously, it would be very interesting to extend the studies carried out with data from other areas both inside and outside medicine. For example, the method described here could be applied to locate outliers in radiodiagnostic images in order to detect tumours. Outside of the medical domain, it could be used to identify outlier credit card transaction sequences to reveal possible card theft fraud. The first example uses image data, whereas the second could be fed by transactional time series.

As already mentioned, the only condition for method application in real environments is basically that a distance measure can be defined between each pair of objects. Distance measure definition is possible provided objects can be characterized (identifying a series of features). This would appear to practicable in any domain that generates data. The only drawback for scaling the method to real commercial environments (for example, a hospital or a bank) is the computational complexity of the clustering process. However, as mentioned above, companies around the world are systematically using segmentation techniques, on which ground this drawback should not pose an obstacle to method use.

The above examples show that the proposed method can be applied whenever it is possible to define a distance measure between objects for clustering purposes, irrespective of the data domain and type . Therefore, method use is not confined to time series. Another point worth mentioning is the role played by experts: they must define the natural dispersion of the domain objects beforehand, which is not always a straightforward matter.

## 6. Conclusions and Future Work

Within the KDD process, outlier detection is regarded as a key data processing task that is able to output more representative and better quality models. Outside the KDD field, outlier detection is, in itself, a task that is very useful for spotting fraud, diagnosing diseases, etc.

In this paper, we propose an outlier detection method based on an initial clustering process. The aim behind the proposal outlined in this paper is to overcome the specificity of many of the existing outlier detection techniques that fail to take into account the inherent dispersion of domain objects.

The outlier detection method is based on four criteria that aim to account for how human beings identify outliers within a set of objects after visually analysing the clusters containing those objects. This has an advantage over other clustering-based outlier detection techniques that are founded on a purely numerical analysis of clusters.

The method described in this paper has been satisfactorily tested on data (time series) from the stabilometry domain, a branch of medicine that studies human balance and postural control, and electroencephalography (EEG), which is a neurological exploration used to diagnose nervous system disorders. To validate this proposal, we ran a series of tests comparing the results determined by a physician and domain expert with the outcomes of applying the outlier detection method. There was a good match between the two sets of results. In particular, the observed false positive rate was not greater than 2% in any case, and the false positives trend was found to be asymptotically positive and close to total precision with larger samples.

The experiments conducted also revealed that the proposed method outperformed other proposals, as, the results, using a more or less automated process, were as good as, or in some cases even better than, other techniques that require the definition of rather a lot of domain-specific parameters.

We also carried out a method efficiency study showing that its only limitation is the computational complexity of the clustering processes, which, even though they are costly, are widely used in KDD within academia and industry.

The outlier detection method uses a number of criteria to generate an outlier factor for each population member. The above factors were translated to a particular specification within the method using a series of equations that account for each factor. Despite the good results, one future line of research is to conduct an in-depth study of the influence of each of the above factors on the outcomes, possibly establishing a new factor weighting within the proposed method depending on their influence on the outcomes.

Another important future line to be addressed in the next stage would be to apply the proposed method in other domains where time series are also representative data, for example, the financial world, and particularly the stock market. It would, of course, be interesting to apply the proposed method to other data types like, for example, spatial data, multimedia data (images), document texts, etc. As already mentioned, we would merely have to define a distance metric between objects as a basis for the clustering process, and the remainder of the method would work with the definition of object dispersion for the specific application domain (medicine, finance, etc.).

Finally, we are considering the possibility of implementing the method described here in a broader expert system, for example in the field of medical diagnosis. In this case, the proposed method would be applied as a preliminary preprocessing step for filtering out outliers and then building more precise reference models of different groups (healthy/ill) for disease diagnosis. It goes without saying that the objects to be filtered and used to build such models must contain data (time series or others) that represent the characteristics of the disease to be diagnosed.

**Conflict of interest statement**: The authors declare no conflict of interest.

**Acknowledgments**: The authors would like to thank Rachel Elliott for translating this paper.

## References

Aggarwal, C. C., & Yu, P. S. (2001). Outlier Detection for High Dimensional Data. *Proceedings of the 2001 ACM SIGMOD international conference on Management of data*.

Agrawal, S, & Agrawal, J. (2015). Survey on Anomaly Detection using Data Mining Techniques. *Procedia Computer Science*, 60: pp. 708-713.

Agrawal, R., Faloutsos, C., & Swami, A. (1993). *Efficient Similarity Search In Sequence Databases*, FODO. Evanston, Illinois, USA.

Agrawal, R., Gehrke, J., Gunopulos, D., & Raghavan, P. (1998). Automatic Subspace Clustering of High Dimensional Data for Data Mining Applications. *Proceedings of the 1998 ACM SIGMOD international conference on Management of data*.

Andrzejak, R. G., Lehnertz, K., Mormann, F., Rieke, C., David, P., & Elger, C. E. (2001). Indications of nonlinear deterministic and finite dimensional structures in time series of brain electrical activity: dependence on recording region and brain state. *Physical review. E, Statistical, nonlinear, and soft matter physics*, 64.

Angiulli, F., Basta, S., & Pizzuti, C. (2006). Distance-Based Detection and Prediction of Outliers. *IEEE Transactions on Knowledge and Data Engineering*, Vol. 18, N. 2.

Barigant, P., Merlet, P., Orfait, J., & Tetar, C. (1972). New design of E.L.A. *Statokinesemeter. Agressol*, 13(C): pp. 69-74.

Baron, J. B. (1964). Presentation d'un appareil pour mettre en evidence les desplacements du centre de gravité du corps dans le polygone de sustentation. Applications pratiques, *Arch Malad Profes*., 25, 1-2: pp. 41-49.

Barona, R. (2003). Interés clínico del sistema NedSVE/IBV en el diagnóstico y valoración de las alteraciones del equilibrio. *Revista de Biomecánica del Instituto de Biomecánica de Valencia (IBV)*, Ed. February.

Barry, R.J., Clarke, A. R., & Johnstone, S.J.(2003). A review of electrophysiology in attention-deficit/hyperactivity disorder:1 Qualitative and quantitative electroencephalography 2. Event-related potentials. *Clinical Neurophysiology*, 114, pp. 171-183 and 184-198.

Black, P. E. (2006). *Manhattan distance*. PhD Thesis, University of Westminster, 2009. in Dictionary of Algorithms and Data Structures [online], Paul E. Black ed., U.S. National Institute of Standards and Technology, (accessed November 2018) Available from: http://www.nist.gov/dads/HTML/manhattanDistance.html.

Black, F. O., & Nashner, L. M. (1984). Vestibulo-spinal control differs in patients with reduced versus distorted vestibular function. *Acta Otolaryngol (Stockh)* 406: pp.100-114.

Black, F. O., & Nashner, L. M. (1985). Postural control in four classes of vestibular abnormalities. In: *Vestibular and Visual Control on Posture and Locomotor Equilibrium*. M Igarashi y FO Black (Eds) Karger Publications, New York, pp. 271-281.

Boniver, R. (1994). Posture et posturographie. *Rev Med Liege*. May 1, 49(5): pp. 285-290.

Bowman, C., & Mangham, C. (1989). Clinical use of moving platform posturography. *Seminars in Hearing* 10 (2): pp. 161-171.

Breunig, M., Kriegel, H.-P., Ng, R., & Sander, J. (2000). LOF: Identifying Density-Based Local Outliers. *Proceedings of the 2000 ACM SIGMOD International Conference on Management of Data*, 29(2), pp. 93-104.

Chakraborty, D., Narayanan, V., & Ghosh, A. (2019). Integration of deep feature extraction and ensemble learning for outlier detection. *Pattern Recognition*, Volume 89, Pages 161-171, https://doi.org/10.1016/j.patcog.2019.01.002.

Chan, K., & Fu, A. W. (1999). *Efficient Time Series Matching by Wavelets*, ICDE, pp. 126-133. Sydney-AUS.

Domingues, R., Filippone, M., Michiardi, P., & Zouaoui, J. (2018). A comparative evaluation of outlier detection algorithms: Experiments and analyses. *Pattern Recognition*, Volume 74, 2018, Pages 406-421, https://doi.org/10.1016/j.patcog.2017.09.037.

Ernst, M. & Haesbroeck, G. (2017). Comparison of local outlier detection techniques in spatial multivariate data. *Data Min Knowl Disc* 31: 371. https://doi.org/10.1007/s10618-016-0471-0.

Ester, M., Kriegel, H.P., Sander, J., & Xu, X. (1996). A density-based algorithm for discovering clusters in large spatial databases with noise. *Proceedings of the International Conference on Knowledge Discovery and Data Mining* (KDD´96), pp. 226-231.

Fayyad, U. M., Piatetsky-Shapiro, & G., Smyth, P. (1996). From Data Mining To Knowledge Discovery: An Overview. In *Advances In Knowledge Discovery And Data Mining*, eds. U.M. Fayyad, G. Piatetsky-Shapiro, P. Smyth, and R. Uthurusamy, AAAI Press/The MIT Press, Menlo Park, CA., pp. 1-34.

Hassan, A. R., & Subasi, A. (2017). A decision support system for automated identification of sleep stages from single-channel EEG signals. *Knowledge-Based Systems*, Volume 128, 2017, Pages 115-124, https://doi.org/10.1016/j.knosys.2017.05.005.

Hodge, V. J., & Austin, J. (2004). A Survey of Outlier Detection Methodologies. *Artificial Intelligence Review*, pp.85-126.

Jaccard, P. (1908). *Nouvelles recherches sur la distribution florale*. Bull. Soc. Vaud. Sci. Nat., 44: pp. 223-270.

Jiang, M. F., Tseng, S. S., & Su, C. M. (2001). Two-phase clustering process for outliers detection. *Pattern Recognition Letters*, 22(6-7), pp. 691-700.

Jolliffe, I. T. (1986). *Principal Component Analysis*. Springer, New York.

Knorr, E., & Ng, R. (1998). Algorithms for Mining Distance Based Outliers in Large Databases. *Proceedings of the 24th International Conference on Very Large Data Bases*, pp. 392-403.

Knorr, E., & Ng, R. (1999). Finding Intensional Knowledge of Distance Based Outliers. *Proceedings of the 25th International Conference on Very Large Data Bases*, pp. 211-222.

Kollios, G., Gunopulos, D., Koudas, N., & Berchtold, S. (2003). Efficient Biased Sampling for Approximate Clustering and Outlier Detection in Large Data Sets. *IEEE Transactions on Knowledge and Data Engineering*, 15(5), pp 1170-1187.

Kovalerchuk, B., Vityaev, E., & Ruiz, J.F. (2000). Consistent Knowledge Discovery in Medical; Diagnosis. *IEEE Engineering in Medicine and Biology Magazine*, Vol. 19, No. 4, pp. 26-37.

Lara, J. A. (2011). *Marco de Descubrimiento de Conocimiento para Datos Estructuralmente Complejos con Énfasis en el Análisis de Eventos en Series Temporales*. Technical University of Madrid, PhD Thesis.

Lara, J. A., Lizcano, D., Pérez, A., & Valente, J. P. (2014). A general framework for time series data mining based on event analysis: Application to the medical domains of electroencephalography and stabilometry. *Journal of Biomedical Informatics*, Vol. 14, pp. 185-199.

Lázaro, M., Cuesta, F., León, A., Sánchez, C., Feijoo, R., & Montiel, M. (2005). Valor de la posturografía en ancianos con caídas de repetición. *Med. Clin.*, Barcelona, pp. 124:207-10.

Loureiro, A., Torgo, L., & Soares, C. (2004). Outlier Detection Using Clustering Methods: a data cleaning application. *Proceedings of KDNet Symposium on Knowledge-based Systems for the Public Sector*.

Martín, E., & Barona, R. (2007). Vértigo paroxístico benigno infantil: categorización y comparación con el vertigo posicional paroxístico benigno del adulto. *Acta Otorrinolaringología Española*, 58(7): pp. 296-301.

Neurocom® International. (2004). *Balance Master Operator's Manual v8.2*. www.onbalance.com (accessed February de 2019).

Nguyen, D., Pongchaiyakul, C., Center, J. R., Eisman, J. A., & Nguyen, T. V. (2005). Identification of High-Risk Individuals for Hip Fracture: A 14-Year Prospective Study. *Journal of Bone and Mineral Research*, 20(11).

Povinelli, R. (1999). *Time Series Data Mining: identifying temporal patterns for characterization and prediction of time series*, PhD. Thesis. Milwaukee.

Raiva, V., Wannasetta, W., & Gulsatitporn, S. (2005). Postural stability and dynamic balance in Thai community dwelling adults. *Chula Med J*, 49(3): pp. 129 – 141.

Rama, J., & Pérez, N. (2003). Artículos de Revisión: Pruebas vestibulares y Posturografía. *Revista Médica de la Universidad de Navarra*, vol. 47, nº 4, pp. 21-28.

Ramaswamy, S., Rastogi, R., & Shim, K.. (2000). Efficient Algorithms for Mining Outliers from Large Data Sets. *Proceedings of the 2000 ACM SIGMOD international conference on Management of data*.

Ren, D., Rahal, I., & Perrizo, W. (2004). A Vertical Outlier Detection Algorithm with Clusters as By-product. *Proceedings of the 16th IEEE International Conference on Tools with Artificial Intelligence*, pp. 22-29.

Romberg, M. H. (1853). *Manual of the Nervous Disease of Man*. London, Syndenham Society, pp. 395-401.

Ronda, J. M., Galvañ, B., Monerris, E., & Ballester, F. (2002). Asociación entre Síntomas Clínicos y Resultados de la Posturografía Computerizada Dinámica. *Acta Otorrinolaringología Española*, 53: pp. 252-255.

Salma N., Mai, B., Namuduri, K., Mamun, R., Hashem, Y., Takabi, H., Parde, N., & Nielsen, R. (2018). Using EEG Signal to Analyze IS Decision Making Cognitive Processes. In: Davis F., Riedl R., vom Brocke J., Léger PM., Randolph A. (eds) *Information Systems and Neuroscience. Lecture Notes in Information Systems and Organisation*, vol 25. Springer, Cham.

Sanz, R. (2000). *Test vestibular de autorrotación y posturografía dinámica*. Verteré, 25: pp. 5-15.

Sarle, W. (1987). *Cubic Clustering Criterion*. SAS Technical Report A-108, SAS Institute Inc.

Sinaki, M., Brey, R. H., Hughes, C. A., Larson, D. R., & Kaufman, K. R. (2005). Significant Reduction in Risk of Falls and Back Pain in Osteoporotic-Kyphotic Women Through a Spinal Proprioceptive Extension Exercise Dynamic (SPEED) Program. *Mayo Clin.*, 80(7): pp. 849-855.

Song, D., Chung, F., Wong, J., & Yogendran, S. (2002). The Assessment of Postural Stability After Ambulatory Anesthesia: A Comparison of Desflurane with Propofol. *Anesth. Analg.*

Sorensen, T. (1957). *A method of establishing groups of equal amplitude in plant sociology based on similarity of species and its application to analyses of the vegetation on Danish commons*. Biologiske Skrifter, Kongelige Danske Videnskabernes Selskab, 5 (4): pp. 1-34.

Stefatos, G., & Hamza, A. B. (2007). Cluster PCA for Outliers Detection in High-Dimensional Data. *Proceedings of the 2007 IEEE International Conference on Systems, Man and Cybernetics*, pp. 3961.3966.

Stockwell, C. W. (1981). Posturography. *Otolaryngol Head Neck Surg.*, 89: pp. 333-335.

Takada H. (2019). Stabilometry to Evaluate Severity of Motion Sickness on Displays. In: Takada H., Miyao M., Fateh S. (eds) Stereopsis and Hygiene. *Current Topics in Environmental Health and Preventive Medicine*. Springer, Singapore.

Tan, P., Steinbach, M., & Kumar, V. (2006). *Introduction to Data Mining*. Addison Wesley, New York.

Torgo, L. (2007). Resource-bounded Fraud Detection. In Progress in Artificial Intelligence, 13th Portuguese Conference in Artificial Intelligence.

Torgo, L., Pereira, W., & Soares, C. (2009). Detecting Errors in Foreign Trade Transactions: Dealing with Insufficient Data. *In Progress in Artificial Intelligence*, Proceedings of the 14th Portuguese Conference in Artificial Intelligence.

Tzallas, A.T., Tsipouras, M.G., & Fotiadis, D.I. (2007). Automatic seizure detection based on time-frequency analysis and artificial neural networks. *Computational Intelligence and Neuroscience*, Vol. 7, N. 3, pp.1-13.

Wang, J.-S., & Chiang, J.-C. (2008). A Cluster Validity Measure with Outlier Detection for Support Vector Clustering. *IEEE Transactions on Systems, Man, and Cybernetics*, Part B, 38(1), pp. 78 – 89.

Wolpaw, J.R, Birbaumer, N. Heetderks, W.J., McFarland, D.J., Peckham, P.H., Schalk, G., Donchin, E., Quatrano, L.A., Robinson, C.J. & Vaughan, T.M. (2000). Brain-computer interface technology: A review of the first international meeting. *IEEE Transactions on Rehabilitation Engineering*, 8(2):164-173.

Yamamoto, M., Ishikawa, K., Aoki, M., Mizuta, K., Ito, Y., Asai, M., Shojaku, H., Yamanaka, T., Fujimoto, C., Murofushi, T., & Yoshida, T. (2018). Japanese standard for clinical stabilometry assessment: Current status and future directions, *Auris Nasus Larynx*, Volume 45, Issue 2, Pages 201-206, https://doi.org/10.1016/j.anl.2017.06.006.

Yang, P., & Huang, B. (2008). A Spectral Clustering Algorithm for Outlier Detection. *International Seminar on Future Information Technology and Management Engineering*, pp. 33-36.

Yoon, K.-A., Kwon, O.-S., & Bae, D.-H. (2007). An Approach to Outlier Detection of Software Measurement Data using the K-means Clustering Method. *Proceedings of the 1st International Symposium on Empirical Software Engineering and Measurement*, pp. 443-445.